\documentclass[10pt,twocolumn,letterpaper]{article}
\usepackage[compact]{titlesec}
\titlespacing{\section}{0pt}{*0}{*0}
\titlespacing{\subsection}{0pt}{*0}{*0}
\titlespacing{\subsubsection}{0pt}{2pt}{1pt}
\usepackage{iccv}
\usepackage{times}
\usepackage{epsfig}
\usepackage{graphicx}
\usepackage{amsmath}
\usepackage{amssymb}
\usepackage{multirow}
\usepackage{subcaption}
\usepackage{enumitem}
\usepackage{xcolor,colortbl}
\usepackage[pagebackref=true,breaklinks=true,letterpaper=true,colorlinks,bookmarks=false]{hyperref}

\iccvfinalcopy 

\def\iccvPaperID{806} 

\ificcvfinal\fi

\begin{document}

\title{Situational Fusion of Visual Representation for Visual Navigation}

\author{Bokui Shen$^1$ \quad Danfei Xu$^1$ \quad Yuke Zhu$^1$ \quad Leonidas J. Guibas$^{1,2}$ \quad Li Fei-Fei$^1$ \quad Silvio Savarese$^1$\\
$^1$Stanford University, $^2$Facebook AI Research\\
{\tt\small \{willshen,danfei,yukez,guibas,feifeili,ssilvio\}@cs.stanford.edu}
}

\maketitle
\ificcvfinal\thispagestyle{empty}\fi

\begin{abstract}

A complex visual navigation task puts an agent in different situations which call for a diverse range of visual perception abilities.
For example, to ``go to the nearest chair'', the agent might need to identify a chair in a living room using semantics, follow along a hallway using vanishing point cues, and avoid obstacles using depth.
Therefore, utilizing the appropriate visual perception abilities based on a situational understanding of the visual environment can empower these navigation models in unseen visual environments.
We propose to train an agent to fuse a large set of visual representations that correspond to diverse visual perception abilities. 
To fully utilize each representation, we develop an action-level representation fusion scheme, which predicts an action candidate from each representation and adaptively consolidate these action candidates into the final action. Furthermore, we employ a data-driven inter-task affinity regularization to reduce redundancies and improve generalization. Our approach leads to a significantly improved performance in novel environments over ImageNet-pretrained baseline and other fusion methods.

\end{abstract}

\begin{figure}[t!]
	\centering
	\includegraphics[width=\columnwidth]{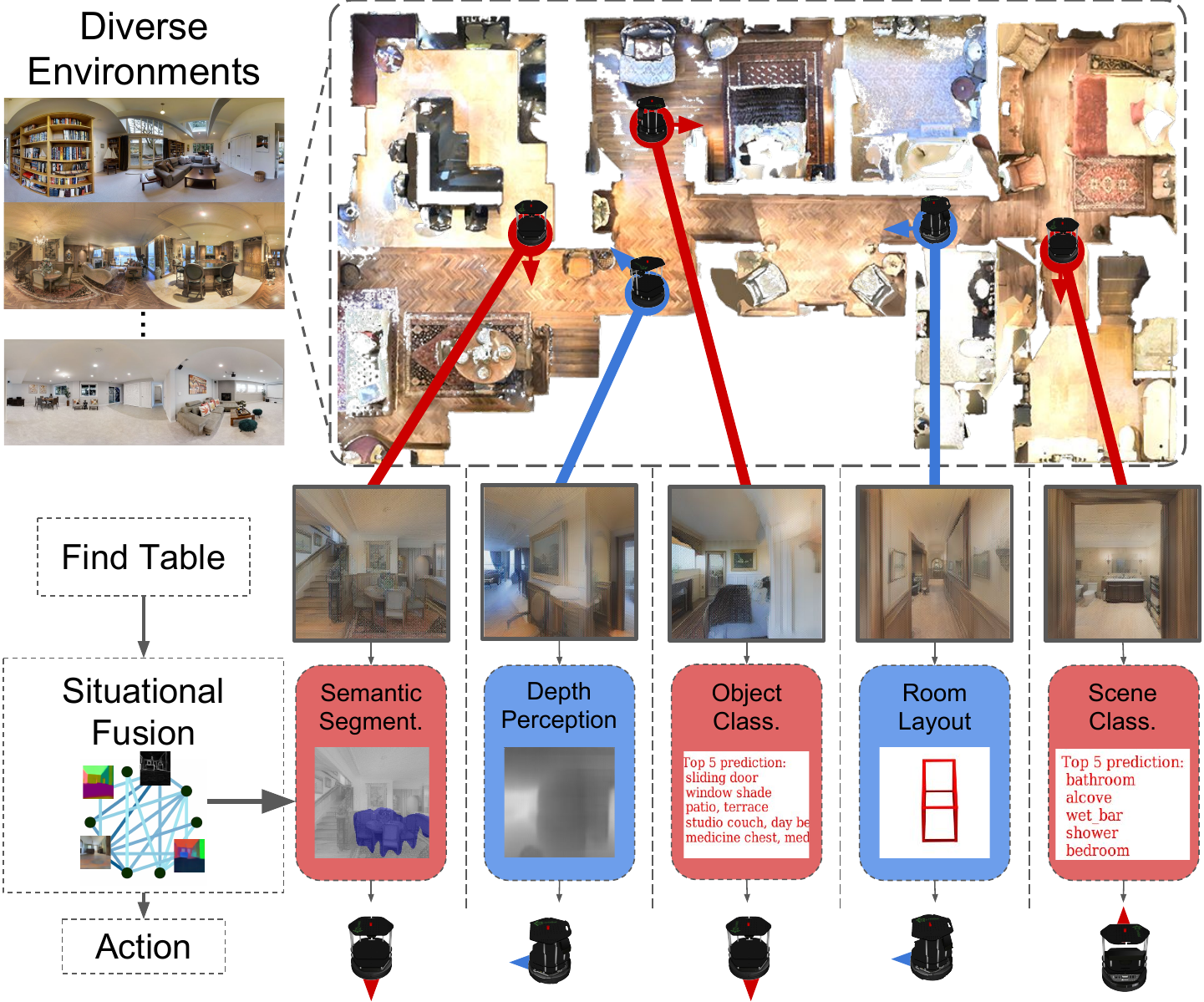}
	\caption{(Top) The visual navigation task requires an agent to handle large diversity in real-world environments. (Bottom) Instead of using a black-box model, we propose to adaptively fuse \emph{visual representations} from a diverse set of vision tasks to better generalize to new environments.}
	\label{fig:pull-figure}
\end{figure}

\section{Introduction}

Assistive robots that can efficiently navigate an everyday home require an extensive repertoire of visual perception abilities. To ``find the nearest cup'', the robot needs a situational combination of different perception abilities. It needs to recognize a cup using object detection, and if no cup is present, identify and navigate to a room that may contain cups using a combination of scene semantic knowledge and geometric cues such as vanishing point and depth. Performing such a complex task in \emph{any home} that may have completely different spatial layouts and appearances demands not only the generalizability of the individual visual perception modules but also the situational combination thereof.

Recently, deep reinforcement learning has made promising progress in goal-directed visual navigation with end-to-end learning ~\cite{levine2016end,zhu2017target}. The guiding principle of imposing \emph{minimal structure} to the learning algorithm and the practically unlimited synthetic training data have warranted the wide applicability of these methods. However, obviating structures and prior knowledge about visual perception can often lead to solutions that overfit to an environment. 
In fact, such success often reckons on training and evaluating in identical or similar environments~\cite{levine2016end,zhu2017target}. Being able to perform complex navigation tasks and generalize to environments with drastically varying appearances as aforementioned is still beyond reach. 

Meanwhile, \emph{visual representations} extracted from a wide range of vision tasks~\cite{Deng2009ImageNetAL,Krishna2016VisualGC} have been greatly effective in generalizing towards new tasks and new domains. It has been a common practice to reuse visual representations trained for standard vision tasks, such as image classification~\cite{krizhevsky2012imagenet}, in new datasets and new problems~\cite{Girshick2015FastR,simonyan2014very}. In this work, we endow visual navigation models with structures and priors based on a diverse set of visual representations. This leads to stronger generalization without losing the wide applicability of end-to-end learning. Our idea is motivated by the recent work of Taskonomy~\cite{zamir2018taskonomy}, which introduced a computational framework for studying the similarity and transferability among vision tasks. 
Their data-driven result shows that effective transfer of representation can happen between tasks of high affinity, which can be used to identify redundancies among tasks and significantly reduce total supervision.
The repertoire of visual representations and their affinity measures from Taskonomy offer a basis for our model to learn generalizable navigation policies that transfer to unseen environments.

Thus, in order to harness visual representations for visual navigation, the primary challenge is to devise a scheme that adaptively fuses the representations based on a situational understanding of the task, while not overfitting to spurious dependencies among visual representations. 
To this end, we introduce an end-to-end approach that fuses visual representations at the action level and uses task affinity as regularization (see Fig.~\ref{fig:pull-figure}).
 We learn an action predictor for each representation, and then combine the action predictions from all representations into the final action output. We also use the data-driven task affinity discovered in Taskonomy as a source of regularization to penalize selection of redundant representations. This method ensures that the fused representation strikes a balance between being informative of the trained task and being generalizable towards new scenarios.

Prior works that addressed generalization of navigation policy \cite{hong2018virtual,muller2018driving} have typically focused on simulation-to-real transfer for low-level motion control tasks. We instead evaluate our approach in visual navigation tasks with higher complexity in the Gibson simulated environments~\cite{xia2018gibson}, which were shown to transfer to the real world without further supervision~\cite{meng2019neural,kang2019generalization}. Our policy takes as input RGB images and produces the action command for the robot. Our results indicate that the use of representations has led to substantially better generalization for a high-level navigation task in unseen environments, and fusion at the action level leads to better generalization and robustness towards noise and subsystem failure. Furthermore, the inter-task affinity regularization promotes the fusion method to select more complementary and less redundant representations.

To summarize, our contributions are twofold: 1) we propose to use visual representations from a diverse set of vision tasks as a prior to learn generalizable policies for visual navigation, and 2) we develop action-level representation fusion and regularization techniques to achieve strong zero-shot generalization results in unseen environments.
\section{Related Work}

\noindent 
\textbf{Representation Learning.} 
Prior work has developed various methods to learn visual representations with deep neural networks using  supervised learning~\cite{Deng2009ImageNetAL}, unsupervised learning~\cite{doersch2015unsupervised,pathakCVPR16context}, and weakly supervised learning~\cite{joulin2016learning} objectives. 
Motivated by the success of using deep features in transfer learning, a series of work has developed visualization and analysis techniques to understand the characteristics of deep features~\cite{mahendran15understanding,zeiler2014visualizing} and the factors that affect the efficacy of transfer~\cite{huh2016makes}. Taskonomy~\cite{zamir2018taskonomy} has recently demonstrated that representations learned on 25 distinct vision tasks can transfer between similar tasks. However, it has focused on static image-based tasks. In contrast, we study how to leverage these diverse visual representations to effectively learn generalizable policies for visual navigation.

\begin{figure*}[t]
\centering
\begin{subfigure}[t]{.25\linewidth}
    \centering
    \includegraphics[width=1.0\linewidth]{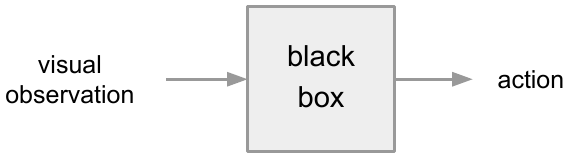}
    \caption{black-box model}
\end{subfigure}
\hspace{.02\linewidth}
\begin{subfigure}[t]{.29\linewidth}
    \centering
    \includegraphics[width=1.0\linewidth]{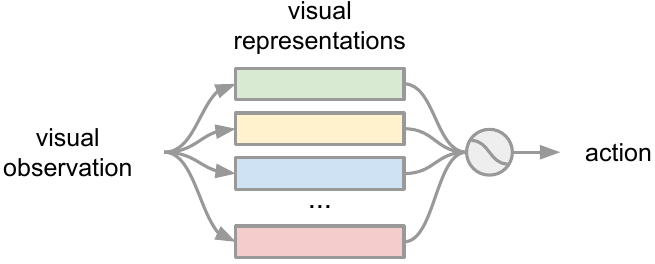}
    \caption{feature-level fusion model}
\end{subfigure}
\hspace{.02\linewidth}
\begin{subfigure}[t]{.33\linewidth}
    \centering
    \includegraphics[width=1.0\linewidth]{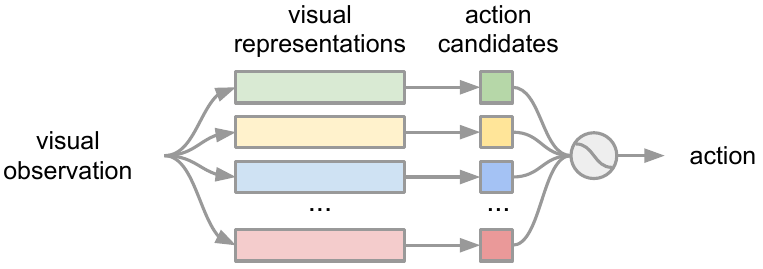}
    \caption{action-level fusion model}
\end{subfigure}
\caption{Three schemes of learning neural network policies from visual perception to action, including end-to-end learning with black-box neural networks, and representation fusion methods at both feature and action levels.}
\label{fig:scheme}
\end{figure*}

\vspace{1mm}

\noindent\textbf{Visual Navigation.} Vision-based navigation for mobile robots has been widely studied in classic robotics literature~\cite{bonin2008visual,thrun2005probabilistic}. The recent boom of deep learning has driven a new wave of development of visual navigation methods that employ the representational power of deep networks. Deep reinforcement learning~\cite{mirowski2018learning,mirowski2016learning,parisotto2017neural,zhu2017target} has been successfully applied to mapless navigation, eliminating the need for an explicit environment map. Other methods have adopted planning modules in the learning pipeline~\cite{gupta2017cognitive}, or tackled the navigation task along with other semantic tasks, such as language grounding~\cite{hermann2017grounded} or visual question answering~\cite{das2017embodied,gordon2018iqa}. Most of the prior work has focused on learning these navigation models end-to-end from scratch rather than reusing visual representations learned from other vision tasks. 
We demonstrate that our model can effectively leverage these representations to achieve stronger generalization when navigating in an unseen environment.

\vspace{1mm}

\noindent\textbf{Generalization in Policy Learning.} A series of methods have been developed to improve the generalization of policy learning towards novel visual observations~\cite{tobin2017domain}, noisy environment dynamics~\cite{mandlekar2017adversarially,rajeswaran2016epopt}, and new task instances~\cite{schaul2015universal,zhu2017target}. Synthetic data has been harnessed as a training source to empower the training of the data-hungry deep network policies. Accordingly, several works have introduced new techniques to close the reality gap, allowing these policies to generalize from simulation to the real world~\cite{hong2018virtual,muller2018driving,zhang2017sim}. In this work, we also use simulated data to train our models. However, our simulator of choice, Gibson~\cite{xia2018gibson},  employs 3D captures of real-world scenes and domain-adapted simulation, which has been shown effective in deploying the simulation-trained policies directly in the real world~\cite{kang2019generalization,meng2019neural}. We focus on improving the generalization of our models across visual scenes with novel ways of fusing the diverse set of visual representations.

\section{Situational Visual Navigation Model}
\label{sec:model}
Our goal is to learn visual navigation policies that generalize better to unseen environments by enabling the agent to adaptively fuse \textit{visual representations} which are trained on a diverse set of vision tasks~\cite{zamir2018taskonomy}.
However, na\"{i}vely fusing representations results in overfitting (Table~\ref{table:succ}), and it is not robust against noise and subsystem failure (Fig.~\ref{fig:drop-figure}).

\begin{figure*}[t]
	\centering
	\includegraphics[width=0.97\textwidth]{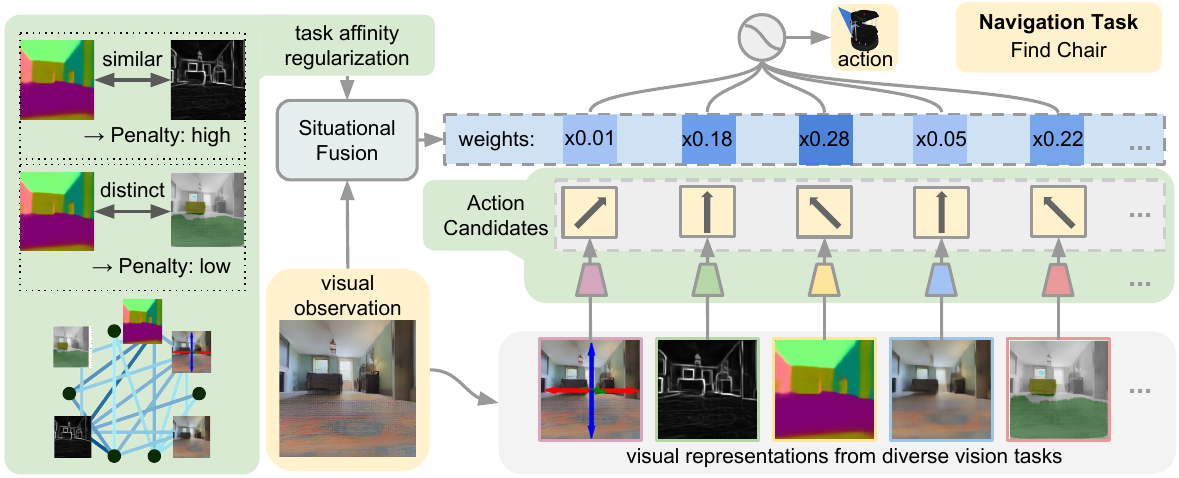}
	\caption{The agent receives an RGB image and its corresponding visual representations. We use a situational-fusion network to adaptively weigh and combine the representations at the action level based on the current observation. The fusion weight is regularized by inter-task affinity to encourage a more balanced selection of the representations for better generalization.}
	\label{fig:model}
\end{figure*}

To address overfitting and to increase model robustness, we propose to combine the representations at the action level (Sec.~\ref{sec:scheme}), which ensures that each representation is individually trained to make meaningful prediction for the overall task.
We also introduce a novel regularization over the fusion weights based on inter-task affinity (Sec.~\ref{sec:task_reg}). It reduces the co-selection of redundant representations and diversifies the use of representations based on the characteristics of each state. Results show that the synergy of action-level fusion and inter-task affinity regularization (Fig.~\ref{fig:model}) yields big performance gain over baselines.

\subsection{Problem Formulation}
\label{sec:setup}
We focus on the task of visual navigation in indoor environments. The agent perceives the environment through its on-board RGB camera~\cite{xia2018gibson}. Following the setup in prior work~\cite{gupta2017cognitive,zhu2017target}, the actions are defined as high-level commands $a_{x, \theta}$, where $\theta$ is the turning angle and $x$ is the stepping distance. We assume that the environments are discretized into a regular octagonal grid, where the agent uses the high-level actions to traverse on the grid. 

We formulate the learning problem as follows. At the beginning of an episode, the agent is randomly spawned at location $p_0=(x_0,y_0)$ in an environment $\mathcal{E}$. At each step, the agent receives a visual observation in the form of RGB image $o_t=\mathcal{O}(p_t, \mathcal{E})$, where $\mathcal{O}$ is a function which returns the current images at location $p_t$ in $\mathcal{E}$. From each visual observation, a set of visual representations $\mathbf{r}_t(o_t)$ can be computed from deep network models trained on a diverse set of vision tasks, where $\mathbf{r}_t(o_t)=\{\mathbf{r}^{1}_t, \mathbf{r}^{2}_t,... \}$. We want to learn a closed-loop navigation policy $\pi(a_t | o_t,\mathbf{r}_t)$, parameterized by neural networks, to map the RGB image $o_t$ and their visual representations $\mathbf{r}_t$ to the action $a_t$ that commands the robot to navigate in the environment. For each task, we specify a Boolean function that determines if the current location $p_t$ satisfies the goal condition. Our objective is to learn an optimal policy $\pi^{*}$ which reaches a goal location from its current location in minimum number of steps.

\subsection{Representations from Diverse Vision Tasks}
\label{sec:taskonomy_bank}

The visual navigation task requires a variety of visual perception abilities, thus demanding visual representations learned from a diverse set of  vision tasks. We leverage the representations from Taskonomy~\cite{zamir2018taskonomy}, which trains deep network models for 25 distinct computer vision tasks at multiple levels of abstraction. For each task, a neural network model is trained with supervised learning to map the input RGB image into a compact representation $\mathbf{r}\in \mathbb{R}^{16\times 16\times8}$, which capsules the information required to solve the task. It has been shown that these representations, being compact in size, are easily transferable to other visual understanding tasks. The compactness and richness of these representations make them appealing for representing the visual perception abilities required in interactive visual tasks. Therefore, we use these 25 representations as a basis to learn our vision-based navigation policies.

Taskonomy also introduced a data-driven affinity score between each pair of the 25 vision tasks. The affinity scores estimate the correlation and redundancy of the representations across these tasks. We demonstrate in Sec.~\ref{sec:task_reg} that we can incorporate such affinity of representations into the process of situational fusion. This encourages the model to make a more balanced selection of representations and reduce overfitting to redundancies.

\subsection{Situational Representation Fusion}
\label{sec:adapt}

A na\"{i}ve approach to learn the navigation policies is to consider deep network as a black-box model and train it with end-to-end learning (Fig.~\ref{fig:scheme}a). While this approach can directly optimize the learning objective from pixel to control, it tends to capture spurious dependencies from limited training data, hindering its generalization ability. Prior work~\cite{zamir2018taskonomy} has proposed to leverage Taskonomy representations via a simple concatenation of $\mathbf{r}_t(o_t)$. However, it offers marginal performance gain over the black-box model (Table~\ref{table:succ}). We hypothesize that this is due to the lack of situational use of representations.

Our key intuition is that different stages of the visual navigation task would require different visual perception skills. For instance, localizing the target object would require semantic understanding, and avoiding obstacles would require geometric reasoning. To satisfy such requirements, the model has to develop a situational understanding of its current perception for decision making. Thus, we introduce a situational-fusion module that adaptively weighs and combines the representations based on the current state. 

A widely used paradigm of combining representations~\cite{dauphin2017language,shazeer2017outrageously} is to use fused representations as joint input to neural network layers (i.e., feature-level fusion illustrated in Fig.~\ref{fig:scheme}b). However, as prior work~\cite{shazeer2017outrageously} pointed out, such fusion method tends to bias towards a few dominating representations and is prone to poor generalization performance. This leads to the two challenges of overfitting and lack of robustness, as highlighted at the beginning of this section. We introduce two effective techniques to deal with the challenges as follows.

\subsubsection{Fusion at the Action Level}
\label{sec:scheme}
\vspace{0.5mm}
As described at the beginning of this section, feature-level fusion faces the challenge of overfitting, and lack of robustness towards noise and subsystem failure.

In robotics, to perform a complex task, compositionality of behaviors~\cite{brooks1986robust, sutton2011horde} promotes to learn a valid behavior for every sub-module, which ensures that all sub-modules are well trained. Then the simpler behaviors are composed at decision-output level into more complex behaviors. Since each sub-module develops a valid behavior, the overall system also becomes more robust towards sub-module errors. 

Inspired by this approach, we propose an alternative way of fusing the representations at the action level (Fig.~\ref{fig:scheme}c). We learn a valid action-prediction model out of each representation, which ensures that all representations are well trained (Fig.~\ref{sec:model}).
Formally, at any time step $t$, as the agent receives RGB image input $o_t$, it uses a situational-fusion network $f$ to output an $n$-dimensional vector $\mathbf{h}_t= f(\mathbf{o}_t)$ for the $n$ representations. $\mathbf{h}_t$ is then normalized with a softmax function to obtain the fusion weight $\mathbf{g}_t= \texttt{softmax}(\mathbf{h}_t)$. We use pretrained Taskonomy modules to compute the $n$ representations for the image $\{\mathbf{r}^1_t, \mathbf{r}^2_t, \ldots, \mathbf{r}^n_t\}$. For each representation $\mathbf{r}^i_t$, an action-prediction module $\pi'_{i,\theta}(a|\mathbf{r}^i_t)$ is independently trained and produces an action candidate $\Tilde{a}^i_t$, thus not suffering from under-training. Each action candidate can be independently taken as a final action for execution, but better decision can be made by using situational-fusion weights $\mathbf{g_t}$ to reweigh and combine the action candidates into the final action output $a_t$:
\setlength{\abovedisplayskip}{4pt}%
\setlength{\belowdisplayskip}{4pt}%
\setlength{\abovedisplayshortskip}{0pt}%
\setlength{\belowdisplayshortskip}{0pt}%
\begin{align*}
a_t &= \sum\limits_{i=1\ldots n} \mathbf{g}^i_t\ \pi'_{i,\theta}(\mathbf{r}^i_t),\  \mathbf{g}_t= \texttt{softmax}(f(\mathbf{o}_t))
\end{align*}
We observe that using this new fusion scheme significantly improves generalization performance over na\"{i}ve fusion scheme (Table~\ref{table:succ}), and the action candidate of each branch achieves reasonable individual performance (Fig.~\ref{fig:branches}). At the same time, it also offers more robustness towards noise and sub-system failure (Sec.~\ref{sec:mod-ana}). However, although individual branches are well developed in the action-level fusion scheme, we observe that the fusion weights still show strong bias towards a few dominating action candidates during inference. In the following section, we propose a novel regularization loss using inter-task affinity.

\subsubsection{Inter-task Affinity Regularization}
\label{sec:task_reg}
\vspace{0.5mm}

The bias of fusion weights towards a few dominating representations makes the overall model prone to overfitting. To deal with the problem, prior work \cite{shazeer2017outrageously} adopted a load-balancing loss (LBL) term to regularize on coefficients of variation $\mathbf{L}_{\texttt{LBL}} = \texttt{CV}(\mathbf{g}_t)$, where $\texttt{CV}$ stands for the coefficients of variation operator. However, LBL loss promotes uniform fusion weights across all representations, and we observe that adding this regularizing term alone to our fusion model offers limited to no performance improvement for both feature-level and action-level fusion. A clear shortcoming of this technique is that it fails to take into account the correlations among representations. 
High correlation between two representations means that choosing both leads to more information redundancy, which can affect the model's generalization ability. Therefore, we further propose to design a loss that leverages correlations of representations as a prior to minimize information redundancy and balance fusion weights.
Taskonomy~\cite{zamir2018taskonomy} uses a data-driven approach to measure pairwise task-affinity by calculating the performance gain from using one task's representation to transfer-learn the other task. We use Taskonomy's data-driven pairwise task affinity as a surrogate for the correlations among different visual task representations.

Formally, for each time step $t$ and fusion weight $\mathbf{g}_t\in \mathbb{R}^{n}$, we want to encourage large weights in $\mathbf{g}^i_t,\mathbf{g}^j_t$ if task $i$ and $j$ have low task affinity $\texttt{aff}(i, j)$, and smaller weight if task affinity is high. To achieve this, we inject a regularizer for task affinity by adding a bilinear loss term~\cite{hastie2004efficient} between fusion-weight vector and task-affinity matrix (Fig.~\ref{fig:model}):
\begin{align}
\mathbf{L}_{g}(\mathbf{g}_t) = \mathbf{g}_t^T\mathbf{F}\mathbf{g}_t,\ \mathbf{F}_{i,j}=\texttt{aff}(i, j)
\end{align}
For every pair of tasks, we measure the product of the task affinity $\texttt{aff}(i, j)$ and its corresponding fusion weights $\mathbf{g}^i_t,\mathbf{g}^j_t$, and calculate $\mathbf{L}_{g}(\mathbf{g}_t)$ as sum of all such products.
For example, \textit{occlusion edge detection} and \textit{surface-normal estimation} are two tasks with high affinity. If the model puts large weights on both tasks in $\mathbf{g}_t$, then the product between their task affinity and fusion weights will be large, which leads to a larger $\mathbf{L}_{g}(\mathbf{g}_t)$ and the selection will be penalized. 
Thus, regularizing on this loss encourages the situational-fusion network to reduce information redundancy and balance fusion weights.

\section{Experiments}

We evaluate our proposed methods of situational fusion in a set of visual navigation tasks. Throughout the experiments, we measure the generalization aspect of different navigation policies, and examine the effectiveness of action-level fusion and the inter-task affinity regularization.

\subsection{Experiment Setup}

\noindent
\textbf{Experimental Testbed. }{We conduct experiments in GibsonEnv~\cite{xia2018gibson} rendering of Matterport3D assets~\cite{Matterport3D}, which were captured with real-world 3D scans and labeled with semantic ground truth. We use a diverse set of 62 simulated indoor environments that  contain our objects of interest. We focus on the high-level visual navigation planning and map the navigation locations to an octagonal grid. At each time step, the agent receives eight RGB images in the octagonal directions obtained from its on-board 360-degree camera. The action space consists of eight actions that step along the eight directions on the octagonal grid and a stop action. Following \cite{gupta2017cognitive}, we preprocess all traversable locations on the octagonal grid and generate a directed graph $\mathcal{G}_{x,\theta}$ that connects these locations. This enables us to acquire the supervision of optimal actions to train our policies through shortest-path algorithms.}
\definecolor{seaborn}{RGB}{215,230,215}



\begin{table*}
\setlength\tabcolsep{1.8pt}
\begin{tabular}[b]{ |c|c|c|c|cccc|ccc|cccc| }
\hline
 & & & & \multicolumn{4}{c|}{\textbf{Feature-level Fusion}} & \multicolumn{7}{c|}{\textbf{Action-level Fusion}} \\ \cline{5-15}
Tasks &Random &ResNet~\cite{he2016deep} &Concat &- &LBL~\cite{shazeer2017outrageously} &T.Aff~\cite{zamir2018taskonomy} &Both &Top 1 &Top 5 &Maj. &- &LBL~\cite{shazeer2017outrageously} &T.Aff~\cite{zamir2018taskonomy} & Both \\ \hline
\small{Bed} &2.0 &39.6 &33.6 &30.1 &32.4 &34.8 &44.0 &55.0 &53.1 &45.3 &48.8 &50.0 &48.8 & 45.7 \\ \hline
\small{Chair} &4.4 &18.5 &21.7 &13.1 &15.7 &16.3 &21.1 &19.9 &24.6 &30.8 &28.9 &31.2 &36.3 & 34.3 \\ \hline
\small{Table} &3.8 &17.3 &17.1 &17.7 &21.4 &19.7 &26.4 &19.5 &24.2 &37.8 &28.9 &25.7 &35.9 & 39.8 \\ \hline
\small{Door} &2.1 &8.3 &29.8 &31.3 &32.5 &33.1 &33.2 &42.5 &54.6 &50.7 &54.6 &52.3 &55.8 & 56.2 \\ \hline
\cellcolor{seaborn}{Avg.} &\cellcolor{seaborn}{3.0} &\cellcolor{seaborn}{20.9} &\cellcolor{seaborn}{25.6} &\cellcolor{seaborn}{23.1} &\cellcolor{seaborn}{25.5} &\cellcolor{seaborn}{26.0} &\cellcolor{seaborn}{\textbf{31.2}} &\cellcolor{seaborn}{34.2} &\cellcolor{seaborn}{39.1} &\cellcolor{seaborn}{41.1} &\cellcolor{seaborn}{40.3} &\cellcolor{seaborn}{39.8} &\cellcolor{seaborn}{\textbf{44.2}} & \cellcolor{seaborn}{44.0} \\ \hline
\end{tabular}
\caption{Quantitative evaluation (success rate) of visual navigation policies in unseen environments.}
\vspace{-3mm}
\label{table:succ}

\end{table*}
\vspace{1mm}

\noindent
\textbf{Task Setups. }{We follow the formulation of semantic navigation tasks~\cite{gupta2017cognitive}. The agent is commanded to ``go to the closest $X$,'' where $X\in\{chair, table, bed, door\}$. We use the ground truth object annotation from \cite{Matterport3D} to label nodes in the graph $\mathcal{G}_{x,\theta}$ with object categories, and define the optimal action as stepping towards the closest instance of the object class. To ensure the plausibility of finding a solution, we ensure the agent to start in a room where at least one instance of the specified object class exists. In our tasks, the maximum shortest-path distance between the agent's starting location and a target object is 32 steps, and the minimum is 6. This setup requires the agent to learn object appearances through algorithmic supervision and find the same object class (different instances) in novel test environments.}
\vspace{1mm}

\noindent
\textbf{Evaluation Protocol. }
We follow the train/test procedures used in prior work~\cite{gupta2017cognitive}. For each task, we use on average 28 environments for training and 14 for testing. An episode is judged to be successfully completed if the agent, given a maximum of 39 steps, ends in a location within 3 steps from the specified object. During testing, we randomly sample a fixed set of 1024 starting locations in the test environments, and report the success rate of each model.
\vspace{1mm}

\noindent
\textbf{Baselines. }
We first compare our method with three baselines which do not use situational fusion:
\begin{itemize}[noitemsep,topsep=1pt,leftmargin=3mm]
  \item \texttt{Random}: a random walk agent that takes a uniformly random action at each time step;
  \item \texttt{ResNet}: a black-box deep neural network that maps from raw pixel inputs to action labels (see Fig.~\ref{fig:scheme}a). We use ResNet-50 model~\cite{he2016deep} as in Taskonomy~\cite{zamir2018taskonomy} with pretrained-weights from ImageNet\cite{Deng2009ImageNetAL}. Same data augmentation as~\cite{he2016deep} is applied.  
  \item \texttt{Concat}: directly concatenating all representations~\cite{zamir2018taskonomy}. 
\end{itemize}

\noindent We then compare our method with na\"{i}ve situational-fusion baselines and prior work~\cite{shazeer2017outrageously}:
\begin{itemize}[noitemsep,topsep=1pt,leftmargin=3mm]
  \item \texttt{Feature-level Fusion}: na\"{i}ve situational fusion of representations at feature-level (Sec.~\ref{sec:adapt}).
  \item \texttt{LBL}: adding load-balance loss~\cite{shazeer2017outrageously}, examined for both feature-level and action-level fusion models.
\end{itemize}

\noindent We report two additional baselines for action-level fusion:
\begin{itemize}[noitemsep,topsep=1pt,leftmargin=3mm]
  \item \texttt{Maj.}: we perform majority voting among the action candidates and select the top one. This corresponds to a simplified version of our action-fusion method where the fusion weight is set to uniform.
  \item \texttt{Top k}: to understand the impact of increasing number of representations, we perform majority voting on the actions from the branches that have the $k$-highest individual feature success rates (Fig.~\ref{fig:branches}).
\end{itemize}
\vspace{1mm}

\noindent
\textbf{Proposed Models. }
We examine our proposed extensions to feature-level fusion. We explore the effectiveness of fusion at action-level (Sec.~\ref{sec:scheme}), which combines a set of action candidates predicted from individual representations (\texttt{action-level fusion}). Then we investigate inter-task affinity regularization (Sec.~\ref{sec:task_reg}) that discourages redundancies in fusion weights (\texttt{T.Aff}).

All models were trained using ADAM~\cite{kingma2014adam} to optimize for the loss function and trained for 16K iterations with batch size of 256 (64 for \texttt{ResNet} baseline). We decay the learning rate by a factor of 10 every 5k iterations. 

\subsection{Quantitative Evaluation}
\label{sec:qual}
Table~\ref{table:succ} compares among baselines and variations of our proposed model. The table consists of five main columns: random agents, ResNet baselines, concatenating representations, feature-level fusion and action-level fusion. We measure each model's performance on all four navigation tasks and report the average across tasks. The numbers are evaluated in test environments unseen during training.

First, we observe a large performance gap between black-box methods that train on raw pixels and our proposed methods that situationally fuse visual representations at action-level with inter-task affinity regularization. The method achieves a 2$\times$ higher success rate than the state-of-the-art pretrained ResNet model~\cite{he2016deep}.
This indicates that correctly leveraging visual representations has a significant effect in improving generalization of the learned policy.

Second, we see that na\"{i}vely using visual representations results in limited performance increase. Directly concatenating 25 different visual representations only results in 5\% performance increase over the ResNet model, and a vanilla feature-level fusion model performs similarly. As mentioned in Sec.~\ref{sec:model}, two challenges (overfitting and lack of robustness) need to be solved. We can see that adding load-balancing loss (\texttt{LBL})~\cite{shazeer2017outrageously} alone only offers marginal help. 

\begin{figure}
	\centering
	\includegraphics[width=0.8\columnwidth]{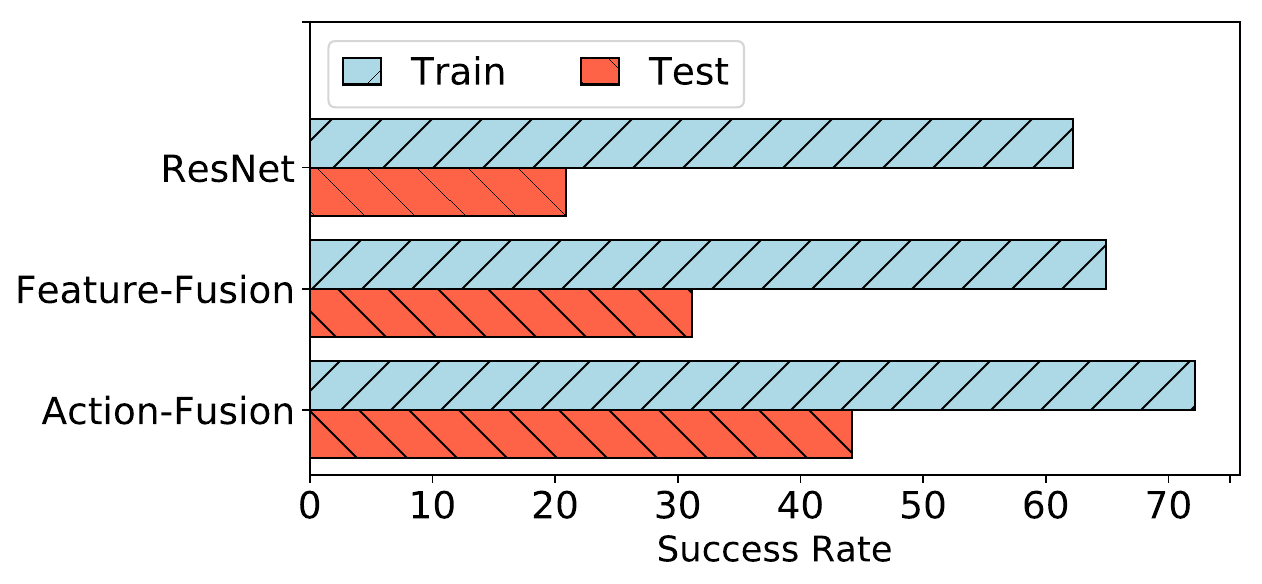}
	\caption{Training and testing performance comparison. Testing scenes are previously unseen environments. Although models achieve similar levels of performance on training scenes, our action-level fusion model generalize significantly better to unseen test environments than baselines.}
	\label{fig:train-test-gen}
\end{figure}

Our proposed action-level fusion explicitly learns an action-prediction model for each representation. Since the load-balancing problem in training is explicitly handled, results show that action-level fusion significantly outperforms \texttt{LBL} loss and that adding \texttt{LBL} loss onto action-level fusion brings no further gain. 
As for comparing with feature-level fusion, action-level fusion shows superior performance. While both schemes have similar training performances, action-level fusion performs much better for testing (Fig.~\ref{fig:train-test-gen}). We hypothesize that action-level fusion suffers less overfitting thanks to its separate handling of each representation. Fusion of the action candidates acts as a model ensemble, which can effectively factor in each representation's contribution in decision making. This hypothesis is further backed by the competitive results of the Majority Voting baseline, which combines the actions with uniform weights. 

\definecolor{myr}{HTML}{D26D6A}
\definecolor{myg}{HTML}{9CC184}
\definecolor{myb}{HTML}{7AA7D7}
\begin{figure*}[t]
	\centering
	\includegraphics[width=0.95\textwidth]{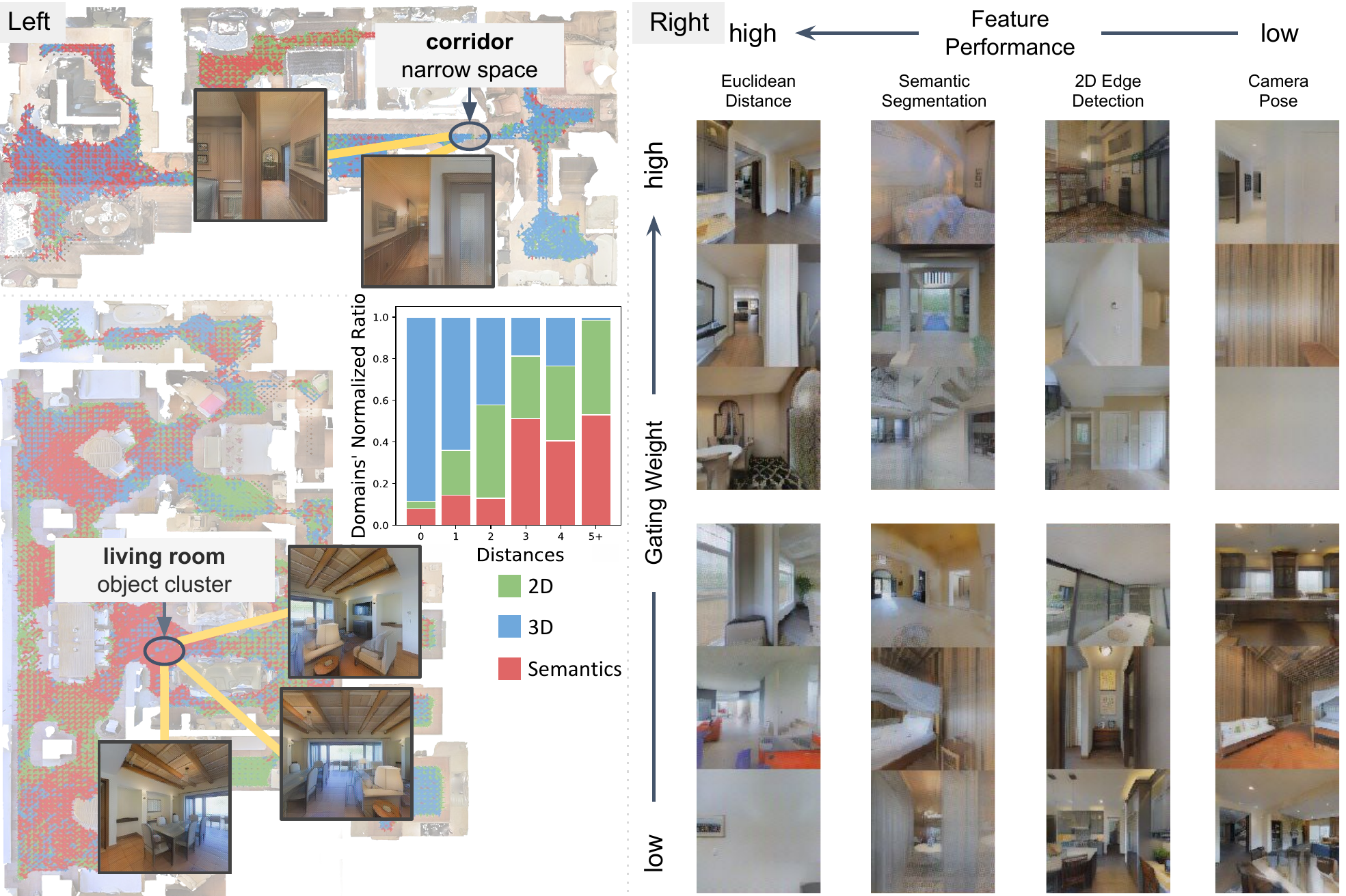}
	\caption{(\textbf{Left}) fusion weight heat-map of different representation domains on two example environments, and a quantitative bar chart of distribution over all testing scenes. 3D representations receive higher weights in narrow space like corridor, while semantic representations become more dominant in object clustered space. (\textbf{Right}) representation's top and bottom activation image based on its fusion weights. Representations with better individual performances were higher weighted in complex scenes and lower in simple scenes; the opposite is true for worse representations.}
	\label{fig:envs}
\end{figure*}

Finally, we see that adding an inter-task affinity regularization (\texttt{T.Aff}) improves the agent's performance in both fusion schemes. \texttt{T.Aff} outperforms \texttt{LBL} loss in both feature-level and action-level fusion. For feature-level fusion, since \texttt{T.Aff} is not designed for dealing with representation under-training, combining it with \texttt{LBL} loss gives a better performance. 
For action-level fusion, which is more effective towards representation under-training than \texttt{LBL}, incorporating \texttt{T.Aff} has achieved the best performance and outperformed the Majority Voting baseline by 3\%. The effect is more significant for tasks of finding table and chair, where the performance boost is over 7\%. Results show that the task affinity regularization is able to reduce spurious dependencies and encourage more balanced selection.

\subsection{Model Analysis}
\label{sec:mod-ana}
\noindent \textbf{Analysis of Fusion Weights: }We conducted qualitative and quantitative studies on the distribution patterns of the situational-fusion weights, results shown in Fig.~\ref{fig:envs}. 

In the first test, we group the representations into three domains: 2D (e.g. autoencoder), 3D (e.g. depth estimation) and semantics (e.g. semantic segmentation). For each location in our environments, we record the fusion weights of our action-level fusion model. We group the weights based on the representation domains, perform normalization, and visualize the top domain, shown in Fig.~\ref{fig:envs} left. The color code is \textcolor{myr}{\textbf{red}} for semantic tasks, \textcolor{myg}{\textbf{green}} for 2D, and \textcolor{myb}{\textbf{blue}} for 3D. 3D representations are the most activated in narrow spaces such as corridor, where the main challenge of the agent is to go through and not collide. In open area with complex object clusters, weight skews more to red, showing more involvement of semantic representations. For quantitative analysis, we use each position's distance (in steps) towards the closest obstacle to its sides as a surrogate for surrounding's openness. For each representation domain, we compute the percentage of positions with highest weights in such domain, and then normalize across distance values. The results are shown in Fig.~\ref{fig:envs}'s bar chart, with Y-axis showing distance and X showing ratio. As surrounding becomes narrower, fusion shifts more from {\color{myr}\textbf{semantics}} to {\color{myb}\textbf{3D}}. 

\begin{figure}
	\centering
	\includegraphics[width=0.95\columnwidth]{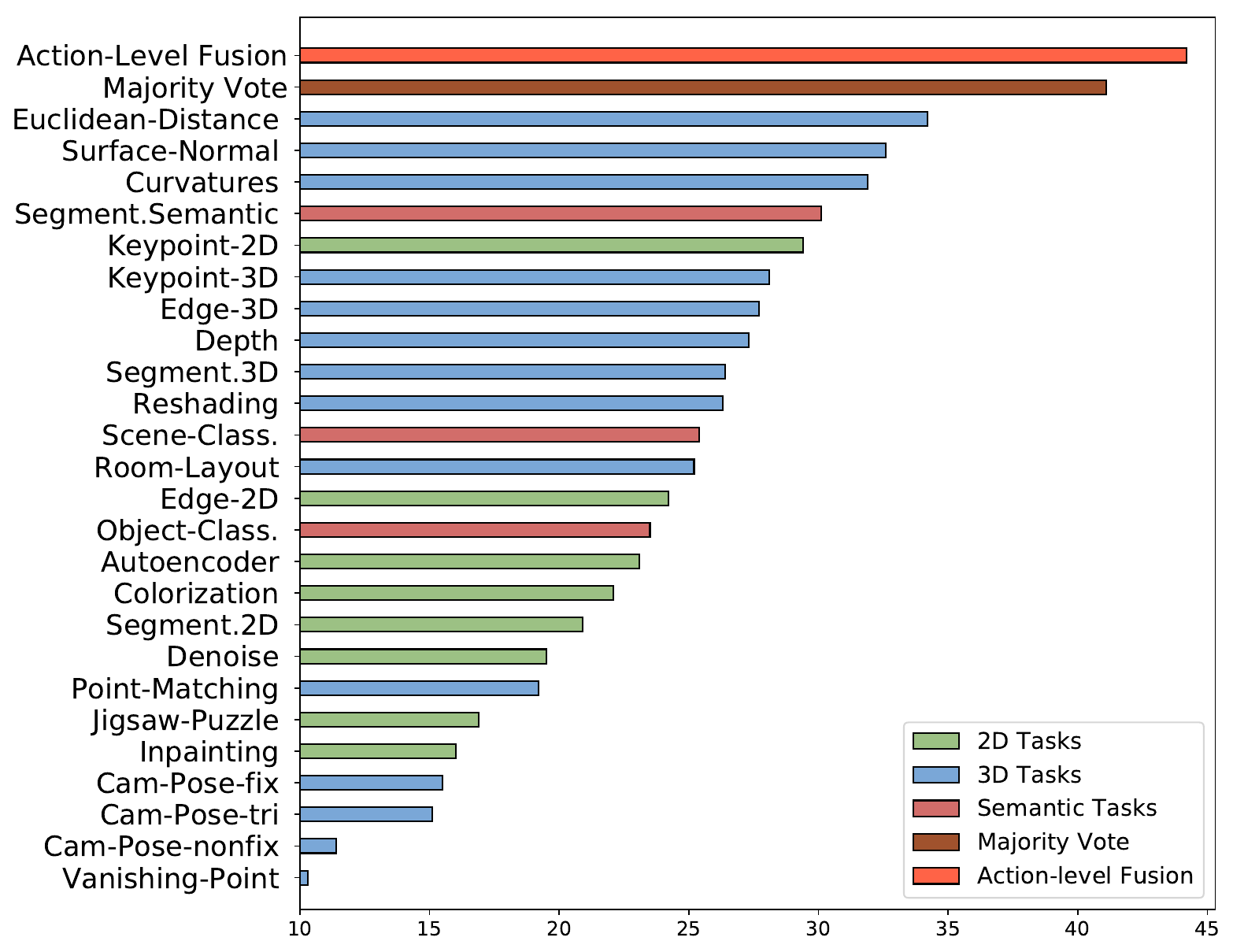}
	\caption{Comparison among our action-level fusion model, majority-voting baseline and each individual representation's performance. Individual branches are color-coded according to task types, with \textcolor{myg}{\textbf{green}} for 2D tasks, \textcolor{myb}{\textbf{blue}} 3D, and \textcolor{myr}{\textbf{red}} semantics.}
	\label{fig:branches}
\end{figure}

In the second test, we examine how weights are distributed between representation branches with different individual performances (see Fig.~\ref{fig:branches}). We sample 100 images from each test environment, and record the fusion weights across samples. Then, for each representation, we select the top 4 images with the highest corresponding fusion weights as well as the lowest bottom 4 images, shown in the right part of Fig.~\ref{fig:envs}. Due to space limit, we spread out only 4 representations of different individual-performance levels.

The columns in Fig.~\ref{fig:envs} are sorted from left to right according to individual representation performances measured in Fig.~\ref{fig:branches}. As we can see, high-performing representations are highly activated when facing more complex scenes with lots of objects and varying spatial layout (Fig.~\ref{fig:envs} top left). They have relatively lower weights when the view becomes more plain (Fig.~\ref{fig:envs} bottom left). Low-performing representations, like 2D edge detection or camera pose estimation, demonstrate the opposite. This suggests that our fusion mechanism learns to put higher reliance on high-performing representations when facing hard decisions (more complex observations), and incorporate low-performing representations to hedge risks when facing simpler decisions.
\vspace{1mm}

\noindent \textbf{Analysis of Individual Representations: }
To investigate the contribution of each visual representation in visual navigation, we quantitatively measure the success rate of directly executing the action candidates from each branch for our action-level fusion models in Fig.~\ref{fig:branches}. 

As we can see, the top three best-performing representations are all 3D geometric tasks. \textit{Euclidean-Distance}, being the top-performing representation, should constantly be consulted so that the agent does not collide into object. \textit{Surface-Normal Estimation}, proved to be one of the best representation to transfer out~\cite{zamir2018taskonomy}, clearly demonstrates its usefulness in navigation tasks. Semantic representations, especially semantic segmentation, ranks relatively high. For our navigation tasks, semantic information is important in an agent's success of locating the target objects. 

Low-level vision tasks, such as vanishing-point estimation and camera-pose estimation, perform poorly as an individual model. These representations have very high abstraction level, and it is hard for the agent to read out scene layout information necessary for complex behaviors.
\vspace{1mm}

\noindent \textbf{Analysis of Model Robustness: } Robustness to unexpected scenarios is very important for a navigation agent. At any moment, a representation might show out-of-distribution noise, and in extreme cases, there might be sub-system failures (such as bugs or attacks) that prevent the agent from accessing its complete representation set. If the agent is unequipped to handle noise or absence of representation, there might be severe consequences.
\begin{figure}[t!]
	\centering
	\includegraphics[width=\columnwidth]{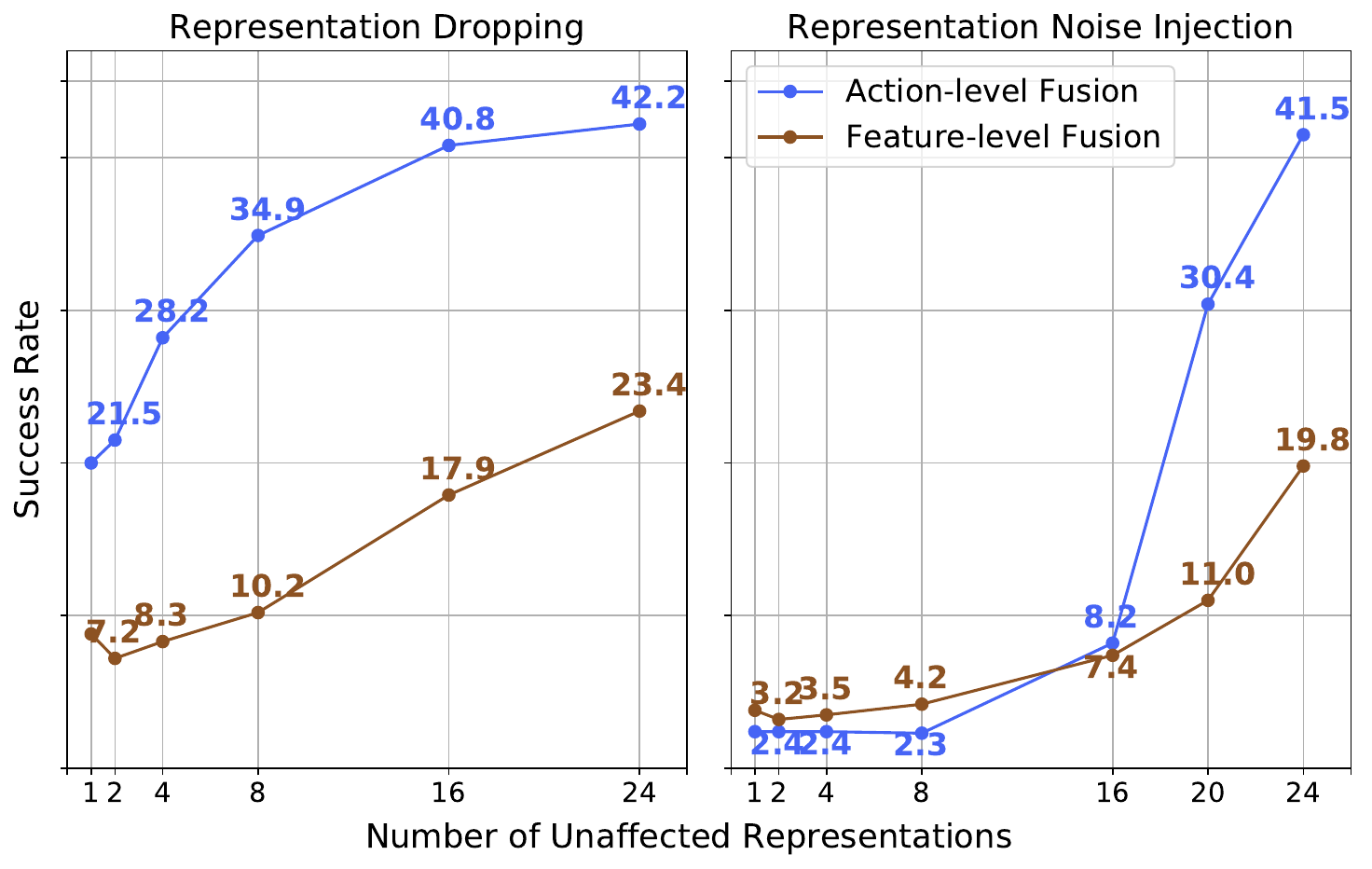}
	\caption{\textbf{Robustness Analysis}:  The left figure shows how performance varies when a number of representations are dropped. The right figure shows how performance varies when a number of representations are set to zero. Horizontal axis shows how many representations remain unaffected.}
	\label{fig:drop-figure}
\end{figure}

Action-level fusion provides a robust way to deal with such challenges. Since each representation individually produces an action output, the downstream effect of missing representations is kept to a minimum. 
On the other hand, any defect to representation will propagate to later layers of the policy network for feature-level fusion. To measure robustness to representation loss, we conduct two tests for the top model of feature-level fusion and of action-level fusion. In both tests, we randomly select a set of representations to be dropped at each step. 
  
For the first test (Fig.~\ref{fig:drop-figure} left), we assume that the loss of representation is noticed by the agent, and it skips the given representation branch by setting the lost representations' fusion weights to zero and normalizing the remaining values. As we see, action-level fusion handles representation loss much better than feature-level fusion.
  
For the second test (Fig.~\ref{fig:drop-figure} right), we randomly replace a set of representations with noise (setting all values of the representations to zero), and the agent treats the affected representation as usual. This is a more challenging setup since the agent might put high weights on the affected representations. We can see that action-level fusion is more robust than feature-level fusion, and is able to have meaningful performance when a random set of 5 representations is affected at each time step.

\vspace{-2mm}

\section{Conclusion}
\vspace{-2mm}

We explored the effectiveness of situationally fusing visual representations from vision tasks to improve zero-shot generalization of an interactive agent in visual navigation tasks. We proposed two novel extensions to fusing representations: action-level fusion and inter-task affinity regularization. Our results suggested that a combination of the two extensions led to better generalization and enhanced robustness of the navigation agent. In a broader scope, these promising results shed light on how to build intelligent systems that can effectively leverage internal representations from other tasks to learn new behaviors. One of the possible future directions is to apply similar principles of representation fusion to other interactive domains, such as active perception and visuomotor learning.

\noindent
\textbf{Acknowledgement: }
We thank Andrey Kurenkov and Ajay Mandlekar for helpful comments. Toyota Research Institute (TRI) provided funds to assist authors with their research, but this article solely reflects the opinions and conclusions of its authors and not TRI or any other Toyota entity.

{\small
\bibliographystyle{ieee_fullname}
\bibliography{egbib}
}

\end{document}